\documentclass[conference]{IEEEtran}
\IEEEoverridecommandlockouts
\usepackage{xcolor}
\usepackage{cite}
\usepackage[
    colorlinks=true,
    citecolor=blue,
    linkcolor=blue,
    urlcolor=blue
]{hyperref}
\usepackage{amsmath,amssymb,amsfonts}
\usepackage{algorithmic}
\usepackage{textcomp}
\usepackage{url}
\usepackage{multirow}
\usepackage{graphicx}
\usepackage{makecell}

\usepackage[table]{xcolor}
\usepackage{booktabs}


\usepackage{tikz}
\usetikzlibrary{positioning}
\usepackage{pgfplots}
\pgfplotsset{compat=1.18}

\def\BibTeX{{\rm B\kern-.05em{\sc i\kern-.025em b}\kern-.08em
    T\kern-.1667em\lower.7ex\hbox{E}\kern-.125emX}}
\begin{document}

% \title{Is Weight Tying Still Beneficial for Modern LLMs in Private Settings Under DP-SGD?\\}
\title{Is Weight Tying Still Beneficial for Decoder-Only LLMs in Private Settings Under DP-SGD?\\}
%%%When DP-SGD Meets Modern LLMs: Do the Benefits of Weight Tying Persist Under DP-SGD in LLMs?\\
\author{
\begin{tabular}{c}
{\small
\textbf{
Razan El Mais,
Ali Chehab,
Ibrahim Issa,
Razane Tajeddine
}
}
\\[0.3em]
\textit{Department of Electrical and Computer Engineering}\\
[0.3em]

\textit{American University of Beirut}\\
Beirut, Lebanon\\[0.5em]
\small
\textbf{\{rre30, chehab, ibrahim.issa, razane.tajeddine\}@aub.edu.lb}\\
\end{tabular}
}


\maketitle

% \textcolor{red}{-Removed experiment of corrected ghost in all results.\\}
% \textcolor{red}{-Stated clearly that although the corrected formulation for ghost norm restores mathematical consistency under weight tying, it no longer preserves the core computational advantage of ghost clipping. Specifically, the cross-term requires explicit materialization, preventing the efficient norm computation trick that ghost clipping relies on.\\}
% \textcolor{red}{-New tables 4,5 and 6 with two new figures: 2 and 3}

% \textcolor{red}{-All sections are edited accordingly}
% \textcolor{red}{-Appendix is edited and enhanced.}

\begin{abstract}
Differentially Private Stochastic Gradient Descent (DP-SGD) is a leading approach for privacy-preserving fine-tuning of large language models (LLMs). Many decoder-only LLMs employ weight tying between input and output embeddings, a design choice originally introduced for parameter efficiency and improved language modeling performance in the non-private setting. However, the impact of weight tying under differentially private training remains largely unexplored. In this work, we investigate the role of weight tying in the DP setting using GPT2 and DistilGPT2 as representative decoder-only architectures. Interestingly, we find that untied embeddings consistently outperform weight-tied models under DP-SGD, achieving gains of up to 4.74\% points in accuracy on SST-2, QNLI, and QQP. Beyond improved utility, untying embeddings enables the use of memory-efficient ghost clipping for DP-SGD. By contrast, weight tying introduces shared-parameter interactions that complicate standard ghost norm computation and largely negate its computational advantages. As a result, untied models achieve over 60\% lower memory usage while preserving the benefits of ghost clipping. Our results indicate that untied embeddings provide a more effective and scalable design for differentially private training of decoder-only LLMs and highlight the need to revisit standard LLM architectural choices in the privacy-preserving setting.

\end{abstract}


%------------------------------------------
\section{Introduction}
Large Language Models (LLMs) have rapidly emerged as foundational components of modern digital infrastructure, with transformative applications spanning healthcare, finance, education\cite{b1,b2}. Advances in model scale, training methodologies, and architectural design \cite{b3,b4,b5} have further accelerated their deployment, enabling increasingly capable LLM-based agents that support a wide range of communication, reasoning, and decision-assistance tasks \cite{b6,b7,b8,b9}. As these models become deeply integrated into user-facing and data-sensitive applications, ensuring the privacy of training data has become a critical challenge.

A growing body of research has demonstrated that LLMs can memorize and unintentionally reveal sensitive information contained in their training corpora through model outputs \cite{b10,b11,b12}. To mitigate this risk, differential privacy (DP) has emerged as one of the most principled frameworks for privacy-preserving machine learning to provide formal privacy guarantees. In particular, DP-SGD \cite{b13} works by clipping per-example gradients and injecting calibrated noise during training. However, applying DP-SGD to modern Transformer-based LLMs remains computationally demanding because per-example gradient clipping incurs substantial memory and runtime overhead. Recent techniques such as ghost clipping \cite{b17} alleviate this bottleneck by computing clipping norms without explicitly materializing individual gradients, significantly improving the scalability of DP training for large models.

A sequence of works has progressively improved the efficiency of per-example gradient computation across various neural architectures spanning fully connected networks, convolutional architectures, and reweighted-loss formulations \cite{b14,b13,b15,b16}.
Building on these advances, ghost clipping \cite{b17} enables memory-efficient differentially private training of Transformer-based LLMs by avoiding explicit per-example gradient instantiation.
Despite its effectiveness, ghost clipping implicitly relies on an important structural assumption: namely, that gradient contributions across parameter groups remain block-separable and can therefore be independently decomposed when computing per-example gradient norms. While this assumption holds for architectures with disjoint parameter blocks, modern decoder-only LLMs such as GPT-style models commonly employ \textit{weight tying} \cite{b18,b19}, where the input embedding and output projection layers share the same parameter matrix.

Weight tying has become a standard architectural choice in many language models because it reduces the number of trainable parameters and often improves generalization performance in non-private training settings. Despite its widespread adoption, its interaction with differentially private optimization techniques remains poorly understood. In particular, existing work on ghost clipping implicitly assumes parameter independence across layers and does not examine the consequences of parameter sharing introduced by tied embeddings. This raises an important question: does weight tying remain beneficial under DP-SGD, and is it fully compatible with the assumptions underlying ghost clipping?

In this paper, we investigate the interaction between weight tying and DP-SGD for fine-tuning decoder-only language models with tied input and output embeddings.
We consider two representative architectures, GPT2 and DistilGPT2, together with corresponding untied variants, and evaluate them on three GLUE benchmark tasks: SST-2, QNLI, and QQP.  %Through a comprehensive study of GPT-2 and DistilGPT-2 under non-private SGD, DP-SGD with standard clipping, ghost clipping, and corrected ghost clipping, 
Our study spans non-private SGD, DP-SGD with standard clipping, and DP-SGD with ghost clipping, allowing us to analyze the effects of weight tying on both utility and computational efficiency under differential privacy. %Through a systematic comparison of non-private SGD, DP-SGD with standard clipping, and DP-SGD with ghost clipping, we show that the conventional benefits of weight tying do not necessarily extend to the private setting. Moreover, we characterize the interaction between weight tying and ghost clipping, revealing a mismatch between shared embeddings and standard ghost norm computation. 
Our main contributions are as follows:
\begin{itemize}
\item We show that untied input and output embeddings consistently improve utility and optimization stability under DP-SGD. At the same time, untying enables memory-efficient ghost clipping, yielding substantial reductions in memory usage.
\item We characterize ghost clipping in the presence of weight tying. Specifically, we show that shared input-output embeddings violate the implicit block-separability assumption underlying ghost norm computation by introducing an additional cross-term in the per-example gradient norm, and we derive the corresponding exact clipping norm.
%\item We derive a corrected norm formulation that accounts for the shared-parameter interaction introduced by tied embeddings and restores consistency with exact per-example gradient norms, but no longer assume ghost core computation. We further show that, although this correction resolves the theoretical discrepancy, it incurs substantial computational overhead, largely offsetting the scalability benefits that make ghost clipping attractive in the first place.
\end{itemize}

The rest of the paper is organized as follows: section \ref{sec: BgandRW} presents the background and related work. Section \ref{sec: ghost clipping Under Weight Tying} introduces our main finding and provides the corresponding analysis. Section \ref{sec: Experimental Setup} describes the experimental setup and implementation details. Section \ref{sec: results and discussion} presents and discusses the experimental results. Section \ref{sec:limitations_future_work} outlines the limitations of this work and directions for future research. Finally, section \ref{sec:conc} concludes the paper.

%------------------------------------------
%------------------------------------------
\section{Background and Related Work}
\label{sec: BgandRW}
\subsection{Differentially Private Stochastic Gradient Descent}

DP has emerged as a rigorous framework for protecting sensitive information in machine learning and providing mathematical guarantees, formally defined as follows. \cite{b20,b21}.

\textbf{Definition 1 (Differential Privacy).}
A randomized mechanism $\mathcal{M}$ satisfies $(\epsilon,\delta)$-Differential Privacy if for any pair of neighboring datasets $D$ and $D'$ differing in a single training example, and for any measurable subset of outputs $\mathcal{S}$,

\begin{equation}
\Pr[\mathcal{M}(D)\in\mathcal{S}]
\leq
e^{\epsilon}
\Pr[\mathcal{M}(D')\in\mathcal{S}]
+
\delta.
\end{equation}

To train deep neural networks under differential privacy, DP-SGD, introduced by Abadi et al. \cite{b13}, has emerged as the standard optimization framework. % DP-SGD operates by computing per-example gradients, clipping each gradient to a predefined norm bound $C$, and subsequently injecting Gaussian noise before parameter updates. Given a mini-batch $\mathcal{B}$, the clipped gradient for sample $i$ is
DP-SGD operates by clipping per-example gradients to a prescribed norm bound and perturbing their aggregate with Gaussian noise prior to the parameter update. Let $\mathcal{B}$ denote a mini-batch and let
\[
g_i = \nabla_\theta \ell(\theta; x_i)
\]
be the gradient of the loss associated with sample $x_i$ with respect to the model parameters $\theta$. For a clipping threshold $C>0$, the clipped gradient is given by
\begin{equation}
\bar{g}_i = g_i \cdot
\min\left( 1, \frac{C}{\|g_i\|_2} \right).
\end{equation} 
The privatized gradient used to update the model parameters is then computed as
\begin{equation}
\tilde{g}
=
\frac{1}{|\mathcal{B}|}
\left(
\sum_{i\in\mathcal{B}}
\bar{g}_i
+
\mathcal{N}(0,\sigma^2C^2I)
\right),
\end{equation}
where $\sigma$ controls the noise magnitude under the Gaussian mechanism \cite{b20,b24}.

\subsection{Efficient Per-Example Gradient Computation}

Although DP-SGD provides strong theoretical guarantees against memorization and membership inference attacks \cite{b22,b23,b32}, its practical scalability remains challenging for modern Transformer-based LLMs. The primary bottleneck stems from the need to compute and clip per-example gradients, which can incur prohibitive memory and computational costs when naively materialized for each training sample. These challenges become particularly severe in billion-parameter models, motivating extensive research on efficient clipping mechanisms and memory-aware DP optimization strategies \cite{b27,b28,b29,b30}. 

%The computational bottleneck of DP-SGD primarily arises from the need to compute per-example gradients. Naively materializing gradients independently for every training sample leads to prohibitive memory consumption and computational cost, especially for large Transformer architectures.

Goodfellow \cite{b14} first demonstrated that for fully connected layers, per-example gradient norms can be computed efficiently using activation vectors and backpropagated error signals without explicitly instantiating full per-example gradients. This observation established the foundation for scalable DP training methods by exploiting the structure of gradient factorization. Subsequent works extended efficient gradient computation to more complex architectures. Rochette et al. \cite{b15} generalized efficient per-example gradient computation to convolutional neural networks, while Lee and Kifer \cite{b16} introduced fast gradient clipping (FGC), which reformulated clipping through reweighted losses to reduce computational overhead during DP optimization. 

More recently, Li et al. \cite{b17} proposed ghost clipping, a memory-efficient clipping mechanism specifically designed for Transformer-scale language models. Rather than explicitly materializing per-example gradients, ghost clipping computes gradient norms analytically from intermediate activations and backpropagation signals. This significantly reduces GPU memory consumption and enables differentially private fine-tuning of modern LLMs at practical scales.

Despite these advances, existing efficient clipping methods typically assume that per-example gradient norms decompose across parameter groups. As we later demonstrate, this assumption is violated in modern decoder-only LLMs employing tied embeddings.

\subsection{Ghost Clipping Mechanism}

Ghost clipping \cite{b17,b28} avoids the overhead of explicitly materializing per-example gradients that require substantially more memory than standard training by exploiting the structure of neural network gradients. Instead of explicitly constructing the full per-example gradient vector, it computes the per-example gradient norm indirectly from quantities already available during the forward and backward passes. For a linear layer with activations $a_i$ and backpropagated error signals $\delta_i$ corresponding to training sample $i$, the per-example weight gradient can be expressed as an outer product,

\begin{equation}
g_i = \delta_i a_i^{T}.
\end{equation}

Using properties of the Frobenius norm, the corresponding gradient norm can be computed without materializing $g_i$:

\begin{equation}
\|g_i\|_F^2
=
\|\delta_i\|_2^2
\|a_i\|_2^2 .
\end{equation}

Ghost Clipping applies this principle across all trainable layers of the model. The squared norm of the full per-example gradient is obtained by summing the layer-wise contributions,

\begin{equation}
\|g_i\|_2^2
=
\sum_{l=1}^{L}
\|g_i^{(l)}\|_2^2 ,
\end{equation}

where $g_i^{(l)}$ denotes the gradient contribution from layer $l$. Once the per-example norm is obtained, the standard DP-SGD clipping coefficient can be computed as

\begin{equation}
R_i
=
\min
\left(
1,
\frac{C}{\|g_i\|_2}
\right),
\end{equation}

where $C$ is the clipping threshold. The loss is then reweighted by $R_i$, allowing the clipped aggregate gradient to be computed through a second backward pass without explicitly storing individual gradients.

By avoiding per-example gradient materialization, Ghost Clipping dramatically reduces GPU memory consumption while preserving the clipping behavior of DP-SGD. However, its norm computation relies on the assumption that layer-wise gradient contributions can be decomposed independently and summed additively. As we show in the next section, this assumption becomes problematic when model parameters are shared through weight tying.

\vspace{0.2cm}
\subsection{Weight Tying in Decoder-Only LLMs}

Weight tying is a parameter-sharing technique in LLMs where the input embedding matrix and the output projection matrix share the same parameters. The method was independently introduced by Press and Wolf \cite{b18} and Inan et al. \cite{b19} to improve parameter efficiency and generalization in neural language modeling.

Let $E \in \mathbb{R}^{V \times d}$ denote the input embedding matrix, where $V$ is the vocabulary size and $d$ is the hidden dimension. In standard untied language models, the output projection layer maintains an independent matrix $W_{\text{out}} \in \mathbb{R}^{V \times d}$. Under weight tying, the model enforces

\begin{equation}
W_{\text{out}} = E.
\end{equation}
As a result, the same parameter matrix is used both to encode input tokens and to generate output logits.
%thereby sharing a single parameter matrix between both the input and output roles.

Prior work demonstrated that weight tying reduces the total number of trainable parameters while often improving perplexity and reducing overfitting in neural language models \cite{b18,b19}. Consequently, tied embeddings became a standard architectural component in modern decoder-only transformer LLMs, including GPT-style architectures.

From an optimization perspective, however, weight tying fundamentally alters the geometry of gradient updates. Since the same parameter matrix simultaneously participates in both the input embedding path and output prediction path, the resulting gradients become structurally coupled. While this coupling is generally benign in standard non-private optimization, its implications under DP-SGD and efficient clipping mechanisms remain largely unexplored. Understanding this interaction is important because training under DP-SGD relies directly on per-sample gradient norms for clipping and noise calibration, making any architectural factor that modifies gradient geometry a potential determinant of utility and computational efficiency.

\subsection{Ghost Clipping Assumptions}

Ghost clipping \cite{b17} relies on a key structural assumption regarding gradient norm decomposition. Specifically, it assumes that gradients across parameter groups remain block-separable, allowing the total per-example gradient norm to be decomposed additively across independent parameter blocks.

Consider model parameters partitioned into $K$ disjoint groups:

\begin{equation}
\theta
=
[\theta_1,\theta_2,\dots,\theta_K].
\end{equation}

If the parameter groups are independent and occupy disjoint coordinates in the parameter space, the squared per-example gradient norm can be decomposed as

\begin{equation}
\|g_i\|_2^2
=
\sum_{k=1}^{K}
\|g_i^{(k)}\|_2^2, \label{eq:seperable}
\end{equation}
where $g_i^{(k)}$ denotes the gradient contribution corresponding to parameter block $\theta_k$.

This additive norm decomposition forms the mathematical foundation of ghost clipping, since it enables efficient computation of gradient norms independently for each layer without explicitly materializing the full gradient vector.

%------------------------------------------
\section{ghost clipping Under Weight Tying}
\label{sec: ghost clipping Under Weight Tying}

\subsection{Understanding the Weight Tying--Ghost Clipping Conflict}

Ghost clipping achieves its scalability by avoiding explicit per-sample gradient materialization and instead estimating gradient norms through layer-wise decompositions. The validity of this approach relies on additive separability, whereby the squared norm of the full per-sample gradient can be expressed as the sum of squared norms computed independently for each parameter group.

\begin{figure*}[t]
\centering
\includegraphics[width=\textwidth]{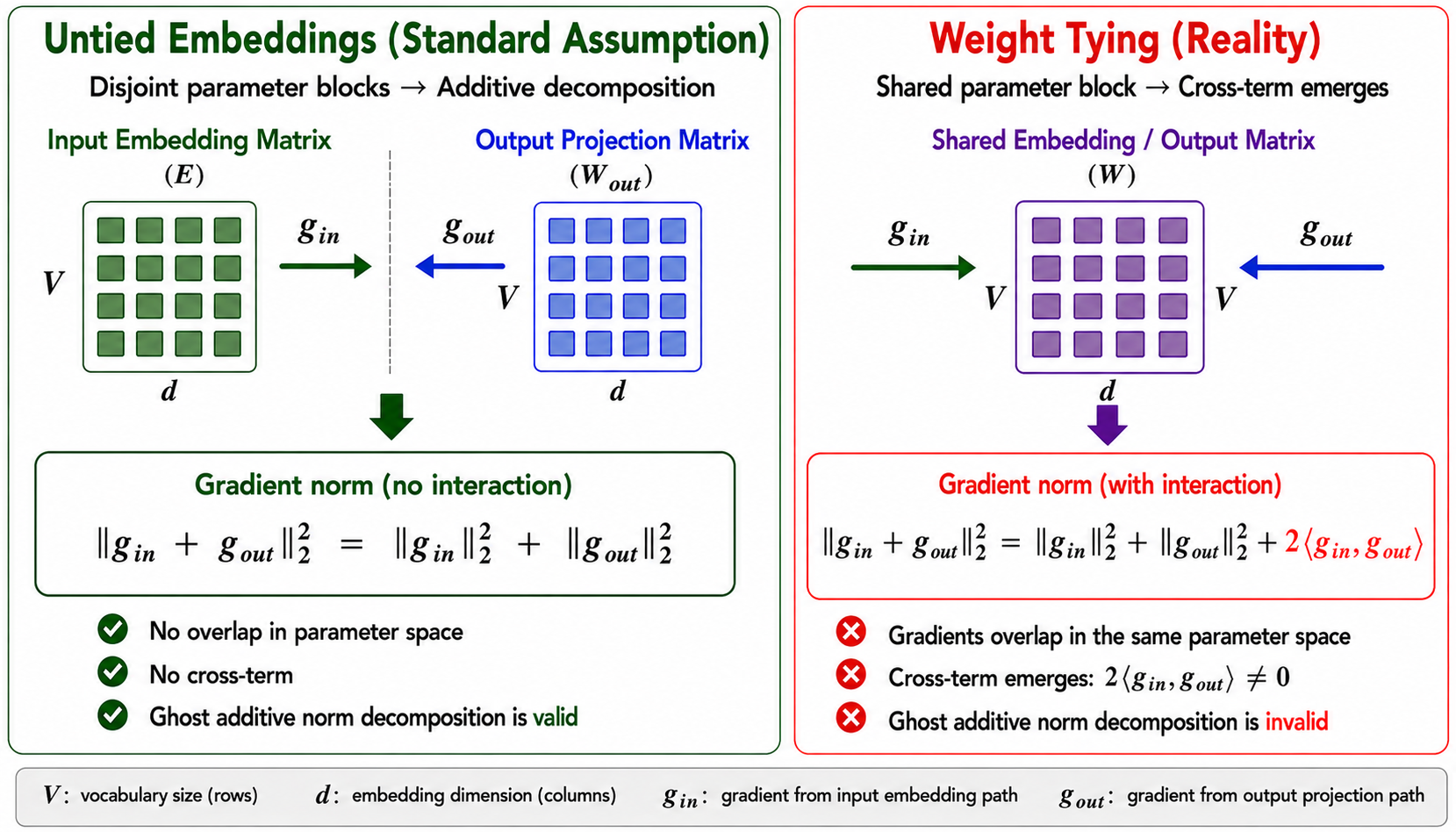}
\caption{The core problem of weight tying that breaks ghost clipping's assumption.}
\label{fig:WT-core-problem}
\end{figure*}

However, modern decoder-only LLMs commonly employ weight tying between the input embedding matrix and the output language-modeling head. Under this design, the same parameter tensor simultaneously serves two computational roles: the input embedding layer and the output projection layer. Consequently, the shared embedding matrix receives gradient contributions from both pathways, as illustrated in Figure~\ref{fig:WT-core-problem}.

Let $g_{\text{in}}$ denote the gradient contribution from the input embedding pathway and $g_{\text{out}}$ denote the contribution from the output projection pathway. The true gradient norm of the shared parameter becomes:

\begin{equation}
\|g_{\text{in}} + g_{\text{out}}\|_2^2
=
\|g_{\text{in}}\|_2^2
+
\|g_{\text{out}}\|_2^2
+
2\langle g_{\text{in}}, g_{\text{out}}\rangle,
\label{eq:cross_term}
\end{equation}
where the additional cross-term captures the interaction between the two gradient components occupying the same parameter coordinates. Consequently,

%The standard ghost clipping formulation does not account for this interaction term because it assumes that each parameter group occupies a distinct set of coordinates in the model parameter vector. Under this assumption, the squared gradient norm can be decomposed into a sum of layer-wise contributions. Weight tying violates this assumption by allowing multiple computational pathways to update the same parameter tensor, thereby introducing the cross-term in Eq.~(\ref{eq:cross_term}). Consequently,

\begin{equation}
\|g_{\text{in}} + g_{\text{out}}\|_2^2
\neq
\|g_{\text{in}}\|_2^2
+
\|g_{\text{out}}\|_2^2,
\end{equation}
violating the assumption in~\eqref{eq:seperable}.

This discrepancy directly affects DP-SGD because clipping coefficients are computed from estimated per-sample gradient norms. When these norms fail to accurately reflect the true gradient magnitude, clipping decisions become distorted, altering optimization dynamics and the effective signal-to-noise ratio during training. Therefore, the discrepancy between weight tying and the standard ghost clipping formulation is not merely an implementation artifact but arises from a mismatch between shared-parameter architectures and the assumptions underlying ghost norm computation.

% \subsection{Ghost Clipping Benefits from Untied Weights}

Our results show that the benefits of weight tying observed in non-private language model training do not necessarily carry over to the differentially private setting. Across both GPT2 and DistilGPT2, untied input and output embeddings consistently achieve a more favorable privacy--utility--efficiency trade-off, yielding higher utility, improved optimization stability, and substantial memory savings under DP-SGD with ghost clipping. To explain this behavior, we investigate the interaction between weight tying and ghost clipping, the state-of-the-art approach for memory-efficient per-sample gradient clipping in Transformer-based models.

%Our analysis reveals that tied embeddings violate the additive norm decomposition assumption implicitly used by standard ghost clipping. Because the embedding matrix participates in both the input and output pathways, the corresponding per-sample gradient norm contains an additional cross-term that is omitted by the conventional ghost norm formulation. As a result,  standard ghost clipping formulation is mathematically incomplete when applied to tied embeddings. 
% Although this discrepancy can be eliminated by deriving and computing the exact norm expression, it no longer reflects the original ghost clipping formulation as the vectors need to be explicitly materialized to compute the cross term. Our experiments show that the resulting computational overhead substantially reduces the efficiency gains that make ghost clipping attractive for large-scale private training. Consequently, the advantages of ghost clipping are most effectively realized when the input and output embeddings are untied. To better understand this phenomenon, we next examine the geometric effects of weight tying on per-sample gradient norms.

\subsection{Corrected Ghost Norm for Tied Embeddings}

To recover the exact per-sample gradient norm under tied embeddings, the interaction term must be explicitly incorporated into the norm computation. The corrected tied-gradient norm is therefore given by:

\begin{equation}
\|g_{\text{tied}}\|_2^2
=
\|g_{\text{in}}\|_2^2
+
\|g_{\text{out}}\|_2^2
+
2\langle g_{\text{in}}, g_{\text{out}}\rangle.
\end{equation}

Unlike the standard ghost approximation, this formulation fully accounts for the interaction between the input embedding and output projection pathways, yielding the exact per-sample gradient norm for the shared embedding matrix. By explicitly incorporating the cross-term, the corrected formulation removes the discrepancy introduced by weight tying and restores consistency with the true gradient norm used by DP-SGD clipping.

However, this correction no longer preserves the core computational advantage of ghost clipping. To the best of our knowledge, no computational method is currently known for evaluating the cross-term without materializing the corresponding gradients. Doing so would partially reintroduce the memory and computational overhead that ghost clipping was designed to eliminate.

Taken together, these findings reveal a trade-off between mathematical correctness and computational efficiency in the presence of weight tying. Untying the input and output embeddings restores additive gradient separability, allowing ghost clipping to operate as originally intended while retaining its scalability advantages.

%------------------------------------------

\section{Experimental Setup}
\label{sec: Experimental Setup}
\subsection{Models}

To study the interaction between ghost clipping and weight tying in decoder-only language models, we evaluate two widely used Transformer architectures: \textbf{DistilGPT2} \cite{b34} and \textbf{GPT2} \cite{b33}. DistilGPT2 is a compact distilled variant containing approximately 82M parameters, while GPT2 contains approximately 124M parameters. Together, these models provide two representative scales for analyzing the effects of weight tying under both private and non-private training.

Both architectures employ causal self-attention and are trained using the standard autoregressive next-token prediction objective. In their original implementations, the input token embedding matrix and output language-modeling head share parameters through weight tying. %a common design choice that reduces model size and is often reported to improve generalization in non-private training.

For each architecture, we evaluate two variants:

\begin{itemize}
\item \textbf{Weight-Tied (WT):} the standard architecture in which the input embedding matrix and output projection layer share parameters.

\item \textbf{Weight-Untied (No-WT):} a modified architecture in which the input embedding matrix and output projection layer are maintained as independent parameter tensors.
\end{itemize}

This controlled setup isolates the effect of embedding parameter sharing, allowing us to systematically evaluate its impact on optimization dynamics, model utility, and ghost clipping behavior under both standard SGD and DP-SGD training.

\subsection{Tasks and Dataset}
All experiments are conducted on three binary classification datasets from the GLUE benchmark: SST-2 (Stanford Sentiment Treebank v2), QNLI (Question Natural Language Inference) and QQP
(Quora Question Pairs). SST-2 is a sentiment analysis task consisting of movie reviews from Rotten Tomatoes labeled as either \textit{positive} or \textit{negative}. QNLI is a question-answering inference task in which question--sentence pairs are classified as \textit{Entailment} or \textit{Not Entailment}. QQP is a paraphrase identification task that classifies pairs of questions as \textit{Duplicate} or \textit{Not Duplicate}.

SST-2 was selected as the primary benchmark because it provides a well-established and computationally efficient setting for studying optimization behavior under both private and non-private training, without the substantial computational overhead associated with large-scale language modeling datasets. QNLI and QQP were additionally included to assess whether the observed effects generalize across different natural language understanding tasks. Dataset statistics are provided in Table \ref{tab:glue-dataset-full} of Appendix \ref{app:datasets}.

Rather than introducing task-specific classification heads, all tasks are formulated as prompt-based autoregressive classification problems that are fully compatible with decoder-only language models. Each input is converted into a textual prompt, and the model is trained to generate a label token corresponding to the target class. For example, SST-2 instances are formatted as:

\begin{quote}
\small
\texttt{Review: <text> Sentiment:}
\end{quote}
with the model generating either \texttt{positive} or \texttt{negative}. Similarly, QNLI and QQP examples are transformed into natural-language prompts, with the model generating tokens corresponding to their respective class labels.

This formulation preserves the standard autoregressive training objective of GPT-style models, allowing the effects of weight tying and ghost clipping to be evaluated without modifying the underlying architecture.

All experiments use a maximum sequence length of 128 tokens. Although evaluation is performed on downstream classification tasks, the primary objective of this work is not to maximize task-specific performance. Rather, our goal is to isolate and analyze the interaction between weight tying and ghost clipping under DP-SGD in a controlled experimental setting while assessing the consistency of the observed behavior across multiple NLP tasks.

\subsection{Training Configurations}
To systematically analyze the interaction between differential privacy, ghost clipping, and weight tying, we evaluate the following training configurations:
\begin{itemize}
\item \textbf{Non-Private SGD}: standard fine-tuning without differential privacy.

\item \textbf{DP-SGD with normal clipping}: conventional differentially private training using explicit per-sample gradient clipping.

\item \textbf{DP-SGD with ghost clipping}: memory-efficient differentially private training using ghost clipping to avoid explicit per-sample gradient materialization.

\end{itemize}
For each training paradigm, we evaluate both weight-tied (WT) and untied (No-WT) embedding configurations whenever applicable. This enables a controlled analysis of the independent and combined effects of DP-SGD, ghost clipping, and embedding parameter sharing.

Hyperparameters were selected through validation-based tuning. For each model and training paradigm, we evaluated multiple combinations of learning rates, clipping norms, batch sizes, and training epochs, selecting the final configuration based on validation accuracy and optimization stability.

For DistilGPT2, We considered learning rates in \{1e-4, 2e-4 and 5e-5\}, clipping norms in \{0.05, 0.1, 0.2 and 0.5\}, batch sizes in {16, 24 and 32}, and training epochs in \{3, 4, 5, 7 and 10\}. The best-performing DP configuration used 10 epochs, Adam optimizer with a learning rate of $10^{-4}$, a batch size of 24, a maximum sequence length of 128 tokens, and a per-example clipping threshold of $C=0.05$. DP-SGD was implemented using the PrivacyEngine from the private-transformers framework.  For GPT2, we considered the same hyperparameter search space and the optimal DP configuration used 4 epochs, a learning rate of $2\times10^{-4}$, a batch size of 24, and a clipping norm of $C=0.5$.
DP-SGD was implemented using the \texttt{PrivacyEngine} from the
\texttt{private-transformers} framework. Privacy loss was tracked using the framework's Rényi Differential Privacy (RDP) accountant. Following the framework's default configuration, the failure probability was set to $\delta=N^{-1.1}$, where $N$ denotes the private training-set size. The Gaussian noise multiplier $\sigma$ was automatically calibrated by the privacy engine according to the target privacy budget, sampling rate, and number of training epochs. Unless
otherwise stated, all main DP experiments used a target privacy budget of $\epsilon=3$. Complete implementation details and experimental scripts are
provided in our public repository. To assess the sensitivity of our findings to this choice, we additionally evaluate DistilGPT2 on SST-2 at
$\epsilon\in\{1,5\}$; these results are reported in Appendix~\ref{sec:epsilon_sensitivity}.

\subsection{Evaluation Metrics}
To evaluate both model utility and the practical impact of ghost clipping under different embedding configurations, we report the following metrics:
\begin{itemize}
\item \textbf{Classification Accuracy}: Mean validation-set classification accuracy across five runs, reported as mean $\pm$ standard deviation.

\item \textbf{GPU Memory Usage}: peak GPU memory consumption during training, measured in terms of allocated and reserved memory. This metric is particularly important for assessing the scalability benefits of ghost clipping.

\item \textbf{Standard deviation Across Random Seeds}: variability in model performance across multiple independent training runs, used as a measure of optimization stability.
\end{itemize}
These metrics capture the key dimensions relevant to differentially private LLM training: predictive utility, computational efficiency, memory scalability, and optimization robustness. Since the central objective of this work is to understand the interaction between ghost clipping and weight tying, evaluating all four dimensions is necessary to characterize the resulting privacy--utility--efficiency trade-offs.

\subsection{Implementation Details}
All experiments were implemented in PyTorch using the HuggingFace Transformers ecosystem together with the \texttt{private-transformers} framework for DP-SGD and ghost clipping support. The implementation includes both standard per-sample clipping and ghost clipping variants.
For tied-weight ghost clipping experiments, we additionally implemented a custom version of the corrected norm formulation derived in Section III. Specifically, the corrected norm explicitly incorporates the interaction term
\begin{equation}
2\langle g_{\text{in}}, g_{\text{out}}\rangle,
\end{equation}
which accounts for the overlap between the input embedding and output projection gradient contributions under weight tying. Unlike the standard ghost formulation, which relies solely on independently computed layer-wise quantities, the corrected implementation requires evaluating interactions between the two shared gradient components.
To compute the corrected norm, the implementation separately evaluates
the input embedding contribution,
the output projection contribution, and
the corresponding interaction cross-term
for each training sample prior to DP clipping. This additional computation enables exact norm estimation for tied embeddings but introduces extra per-sample overhead that reduces the efficiency advantages of ghost clipping.
All experiments were conducted on a single NVIDIA A100 40GB GPU. Runtime, peak GPU memory allocation, and peak GPU reserved memory were recorded throughout training to quantify the computational and memory implications of the different clipping strategies.

%------------------------------------------

\section{Results and Discussion}
\label{sec: results and discussion}

\subsection{Weight Tying in Standard Non-Private Training}
We first examine the effect of weight tying under standard non-private optimization to establish a baseline understanding of its architectural role independent of differential privacy. The results are summarized in Table~\ref{tab:non_dp_wt}.

Across both DistilGPT2 and GPT2, weight tying preserves predictive performance while substantially improving parameter efficiency. For DistilGPT2, tied and untied configurations achieve identical average accuracy of $86.77 \pm 0.76\%$, while weight tying reduces the number of trainable parameters from 120.5M to 81.9M, corresponding to a reduction of approximately $32\%$. Similarly, GPT2 achieves identical average accuracy of $89.09 \pm 0.69\%$ under both configurations while reducing the number of trainable parameters from 163.0M to 124.4M, a reduction of approximately $24\%$.

Weight tying also reduces GPU memory consumption across both models. For DistilGPT2, peak memory usage decreases from 3824.64 MB to 3676.64 MB, while GPT2 exhibits a similar reduction from 5538.99 MB to 5390.99 MB. Runtime differences are comparatively small. DistilGPT2 experiences a modest reduction in training time, whereas GPT2 shows nearly identical runtimes under tied and untied configurations, suggesting that the primary benefit of weight tying stems from parameter and memory efficiency rather than computational speed.

Overall, these results are consistent with prior observations in language modeling literature: weight tying acts as an effective parameter-sharing mechanism that reduces model size and memory requirements without sacrificing predictive performance. Importantly, under standard non-private optimization, weight tying introduces neither optimization instability nor utility degradation. The identical performance of tied and untied configurations further suggests that any differences observed under DP-SGD are unlikely to be driven solely by model capacity, but rather by the interaction between weight sharing and private optimization.

\begin{table*}[!t]
\centering
\caption{Effect of Weight Tying under Standard Non-Private Training on SST-2. Weight tying preserves predictive utility while improving parameter efficiency, reducing GPU memory usage, and slightly shortening runtime across both decoder-only models.}
\label{tab:non_dp_wt}

\renewcommand{\arraystretch}{1.2}

\begin{tabular}{|c|c|c|c|c|}
\hline
\textbf{Model} & \textbf{Setting: SGD} & \textbf{Accuracy (\%)} & \textbf{GPU Memory (MB)} & \textbf{Trainable Params} \\
\hline

\multirow{2}{*}{DistilGPT2}
& WT 
& $86.77 \pm 0.76$
& $\mathbf{3676.64}$
& $\mathbf{81,912,576}$
\\ \cline{2-5}

& No-WT
& $86.77 \pm 0.76$
& $3824.64$
& $120,509,952$
\\
\hline

\multirow{2}{*}{GPT2}
& WT
& $89.09 \pm 0.69$
& $\mathbf{5390.99}$
& $\mathbf{124,439,808}$
\\ \cline{2-5}

& No-WT
& $89.09 \pm 0.69$
& $5538.99$
& $163,037,184$
\\

\hline
\end{tabular}
\end{table*}

\subsection{Weight Tying Under DP-SGD with Normal Clipping}
We next evaluate the effect of weight tying under Differentially Private SGD using conventional per-sample clipping. The results are summarized in Table~\ref{tab:dp_normal_clip}.

In contrast to the non-private setting, weight tying consistently degrades utility under DP optimization. This behavior is likely due to the interaction between per-sample gradient clipping and the shared embedding parameters introduced by weight tying, which amplifies gradient distortion under DP-SGD and increases the impact of injected noise on optimization.
For DistilGPT2, tied embeddings achieve an average accuracy of $80.61 \pm 1.32\%$, whereas untying the embedding and output projection layers improves performance to $83.23 \pm 0.25\%$, corresponding to a gain of approximately $2.6$ percentage points. Similarly, GPT2 improves from $78.44 \pm 1.73\%$ under tied embeddings to $81.48 \pm 1.07\%$ after untying, yielding an improvement of approximately $3.0$ percentage points.

Notably, these gains cannot be attributed solely to increased model capacity. 
As shown in Table~\ref{tab:non_dp_wt}, despite introducing approximately 38.6M additional trainable parameters, the untied variants achieve identical performance to their tied counterparts under standard non-private training for both DistilGPT2 and GPT2. The utility advantage emerges only under DP-SGD, suggesting that the improvement is associated with the interaction between parameter sharing and private optimization rather than increased capacity alone.
%As shown in Table~\ref{tab:non_dp_wt}, tied and untied configurations achieve identical performance under standard non-private training for both DistilGPT2 and GPT2. The utility improvements observed here therefore arise from the interaction between weight sharing and private optimization rather than from additional model parameters.%
Beyond utility, untying substantially improves optimization stability. For DistilGPT2, the standard deviation across random seeds decreases from $1.32$ to $0.25$, while GPT2 exhibits a reduction from $1.73$ to $1.07$. These results suggest that weight tying interacts unfavorably with DP-SGD optimization.

%Although untied models require slightly more memory due to the additional embedding parameters, the runtime differences remain modest relative to the observed improvements in utility and stability.

 %The shared embedding matrix simultaneously receives gradient contributions from both the input embedding and output projection pathways, creating optimization dynamics that differ fundamentally from those observed under standard SGD. In the following sections, we show that this behavior is closely connected to the assumptions underlying ghost clipping and the geometry of per-sample gradient norms under shared-parameter architectures.
\begin{table*}[!t]
\centering
\caption{Effect of Weight Tying under DP-SGD with normal clipping on SST-2. Untying embeddings consistently improves utility and optimization stability under DP across both DistilGPT2 and GPT2.}
\label{tab:dp_normal_clip}
\renewcommand{\arraystretch}{1.25}

\begin{tabular}{|c|c|c|c|}
\hline
\textbf{Model} & \textbf{Setting: DP-SGD with normal clipping} & \textbf{Accuracy (\%)} & \textbf{Peak GPU Memory (MB)} \\
\hline

\multirow{2}{*}{DistilGPT2}
& WT
& $80.61\pm 1.32$
& $13813.55$
\\ \cline{2-4}

& No-WT
& $\mathbf{83.23 \pm 0.25}$
& $14256.41$
\\
\hline

\multirow{2}{*}{GPT2}
& WT
& $78.44 \pm 1.73$
& $18530.09$
\\ \cline{2-4}

& No-WT
& $\mathbf{81.48 \pm 1.07}$
& $18973.67$
\\
\hline
\end{tabular}
\end{table*}

\subsection{Weight Untying Under DP-SGD with Ghost Clipping}
We next evaluate ghost clipping under untied embeddings, where the additive gradient separability assumption remains valid. This experiment serves as a controlled setting for assessing whether ghost clipping can reproduce the improved performance of normal clipping DP-SGD in untied weight setting while providing its intended memory-efficiency benefits.

Across both models, ghost clipping preserves the same predictive utility and optimization stability achieved by standard DP-SGD with normal clipping as shown in Table \ref{tab:dp_ghost_clip_untied}. For DistilGPT2, both methods achieve an identical accuracy of $83.23 \pm 0.25\%$, and for GPT2 they achieve the same accuracy of $81.48 \pm 1.07\%$. These results indicate that ghost clipping introduces no measurable loss in utility when the underlying separability assumptions are satisfied.

The primary benefit of ghost clipping is observed in memory consumption. For DistilGPT2, peak GPU memory usage decreases from approximately $14.3$ GB to $4.7$ GB, corresponding to a reduction of roughly $67\%$. Similarly, GPT2 reduces peak memory consumption from approximately $19.0$ GB to $6.8$ GB, a reduction of more than $64\%$. Runtime remains comparable between the two clipping strategies, with only minor differences observed across models.

These results confirm that ghost clipping functions as intended when the model architecture satisfies the assumptions underlying its layer-wise norm decomposition. Under untied embeddings, ghost clipping reproduces the utility and stability of standard DP-SGD while significantly reducing memory requirements, validating its effectiveness as a scalable mechanism for private training of decoder-only language models.

\begin{table*}[!t]
\centering
\caption{Comparison between ghost clipping and standard DP-SGD normal clipping under untied embeddings (No-WT) on SST-2. When the additive gradient separability assumption remains valid, ghost clipping preserves identical utility and optimization stability while substantially reducing GPU memory consumption across both DistilGPT2 and GPT2.}
\label{tab:dp_ghost_clip_untied}

\renewcommand{\arraystretch}{1.25}

\begin{tabular}{|c|c|c|c|c|}
\hline
\textbf{Model} & \textbf{Clipping Strategy (No WT)} & \textbf{Accuracy (\%)} & \textbf{Peak GPU Memory (MB)} \\
\hline

\multirow{2}{*}{DistilGPT2}
& ghost clipping
& $83.23\pm 0.25$
& $\mathbf{4747.19}$
\\ \cline{2-4}

& normal clipping
& $83.23 \pm 0.25$
& $14256.41$
\\
\hline

\multirow{2}{*}{GPT2}
& ghost clipping
& $81.48 \pm 1.07$
& $\mathbf{6786.90}$
\\ \cline{2-4}

& normal clipping
& $81.48 \pm 1.07$
& $18973.67$
\\

\hline
\end{tabular}
\end{table*}

\vspace{0.2cm}
    \subsection{Weight Tying Under DP-SGD with Corrected Cross-Term Computation}

To determine whether ghost clipping can be made compatible with tied embeddings, we implemented the corrected norm formulation derived in Section \ref{sec: ghost clipping Under Weight Tying}, explicitly incorporating the interaction cross-term introduced by the shared embedding matrix. The corrected formulation restores consistency with standard DP-SGD clipping under weight tying. Across both DistilGPT2 and GPT2, the corrected method achieves identical accuracy to normal clipping DP-SGD with tied embeddings, indicating that the derived norm successfully captures the missing gradient interaction and recovers the correct clipping behavior.

To the best of our knowledge, no computational method is currently known for evaluating the cross-term without materializing the corresponding gradients. As a result, our implementation incurs substantial additional overhead. For DistilGPT2, training time increases from approximately $110.39 \pm 3.55$ minutes under standard DP-SGD to $877.07 \pm 2.92$ minutes under the corrected tied formulation. Similarly, GPT2 requires $607.15 \pm 5.64$ minutes of training time when using the corrected norm.

These results suggest that, while exact clipping under weight tying is achievable through explicit cross-term correction, doing so may substantially reduce the practical efficiency gains associated with ghost clipping.

\vspace{0.2cm}

\subsection{Complete DP Perspective}
Combining the results across all experimental settings reveals a consistent pattern for both DistilGPT2 and GPT2, summarized in Tables~\ref{tab:complete-dp_tradeoff-distligpt2}, \ref{tab:complete-dp_tradeoff-gpt2} and \ref{tab:all_exp_comparison} for SST-2 and Tables \ref{tab:distilgpt2_dp_complete-QNLI}-\ref{tab:gpt2_dp_complete-QQP} of the Appendix \ref{sec: qnliandqqp} for QNLI and QQP respectively.
We additionally evaluate sensitivity to the privacy budget on DistilGPT2 and SST-2 at $\epsilon\in\{1,5\}$, complementing the main
$\epsilon=3$ experiments. As reported in Appendix~\ref{sec:epsilon_sensitivity}, the same qualitative trend persists
across all three privacy budgets: untied embeddings consistently improve
utility over weight tying, while Ghost Clipping preserves the utility of
normal clipping under untied embeddings with substantially lower memory usage.

Under standard non-private optimization, weight tying behaves exactly as intended: it substantially reduces the number of trainable parameters and lowers memory consumption without affecting predictive performance. This observation is consistent with the conventional motivation for weight tying in language models and confirms that tied and untied embeddings exhibit equivalent utility in the absence of differential privacy.

The situation changes markedly under DP-SGD. Across both models, untied embeddings consistently achieve higher accuracy and lower variability across random seeds than their tied counterparts, indicating that the benefits of weight tying do not naturally transfer to the private training regime. Importantly, these gains cannot be explained solely by increased model capacity, since tied and untied configurations achieve identical performance under standard SGD.

Ghost clipping further clarifies this behavior. When embeddings are untied, the additive gradient separability assumption remains valid, allowing ghost clipping to reproduce the utility and stability of standard DP-SGD while reducing GPU memory consumption by more than 60\%. In contrast, weight tying violates this separability assumption by introducing a shared-parameter interaction term into the per-sample gradient norm. Although explicitly accounting for this interaction restores consistency with standard clipping, the resulting computational overhead largely eliminates the scalability benefits that make ghost clipping attractive.

Taken together, these findings reveal that efficient differentially private optimization is not entirely architecture-agnostic. The effectiveness of methods such as ghost clipping depends not only on the optimization algorithm itself but also on architectural properties of the underlying model. In particular, weight tying alters the geometry of per-sample gradients in a manner that conflicts with the assumptions enabling efficient norm computation. Consequently, untied embeddings combined with ghost clipping provide the most favorable privacy--utility--efficiency trade-off among all configurations evaluated in this study.

% 

%_________________%

% %Edited overall figure
% \begin{figure*}[t]
% \centering
% \scriptsize

% \begin{tikzpicture}[
%     node distance=4cm,
%     every node/.style={align=center, font=\small},
%     box/.style={draw, rounded corners, minimum width=4cm, minimum height=2.2cm},
%     arrow/.style={->, thick}
% ]

% % ---------------- Node 1 ----------------
% \node (wt) [box, fill=red!10]
% {\textbf{WT + Normal Clipping}\\[12pt]

% Accuracy: 78.44 $\pm$ 1.73\% \\
% Memory: 18530.09 MB};

% % ---------------- Node 2 ----------------
% \node (untied) [box, right=3cm of wt, fill=orange!12]
% {\textbf{No WT + Normal Clipping}\\
% [12pt]

% Accuracy: \textbf{81.48 $\pm$ 1.07\%} \\
% Memory: 18973.67 MB \\};

% % ---------------- Node 3 ----------------
% \node (ghost) [box, right=3cm of untied, fill=green!12]
% {\textbf{No WT + Ghost Clipping}\\[12pt]

% Accuracy: 81.48 $\pm$ 1.07\% \\
% Memory: \textbf{6786.90 MB} \\};

% % ---------------- Arrows ----------------
% \draw[arrow] (wt) -- node[midway, above] {Untie weights\\ Utility: $\uparrow$ Higher\\
% Memory: $\uparrow$ Higher} (untied);

% \draw[arrow] (untied) -- node[midway, above] {Ghost clipping\\ Utility: preserved\\
% Memory: $\downarrow\downarrow$ Lower} (ghost);

% \end{tikzpicture}

% \caption{Complete DP perspective for GPT2 across weight tying and clipping paradigms on SST-2 dataset. Untied embeddings consistently improve utility and stability under DP-SGD, while ghost clipping preserves these gains with substantially lower GPU memory usage resulting in best overall DP configuration.}
% \label{fig:utility_memory_tradeoff}
% \end{figure*}

%_________________%
\begin{table*}[t]
\centering
\caption{Comparison of weight tying and clipping strategies under DP-SGD on SST-2 (DistilGPT2). The best utility is achieved with untied weights, while the most efficient configuration in terms of memory is Ghost Clipping with untied weights.}
\label{tab:complete-dp_tradeoff-distligpt2}

\renewcommand{\arraystretch}{1.3}
\scriptsize
\begin{tabular}{|p{3.6cm}|c|c|c|}
\hline
\textbf{Configuration} & \textbf{Accuracy (\%)} & \textbf{Memory (MB)}  &
\textbf{Effects}\\
\hline

\textbf{WT + Normal Clipping} 
& 80.61 $\pm$ 1.32 
& 13813.55
& Baseline with moderate memory\\
\hline

\textbf{No WT + Normal Clipping} 
& \textbf{83.23 $\pm$ 0.25} 
& 14256.41 
& Untie weights $\Rightarrow$ \textcolor{green!45!black}{$\uparrow$ Higher} Utility \& \textcolor{red!45!black}{$\uparrow$ Higher} Memory\\
\hline

\textbf{No WT + Ghost Clipping} 
& 83.23 $\pm$ 0.25 
& \textbf{4747.19} 
& Ghost clipping $\Rightarrow$ preserved Utility  \& \textcolor{green!45!black}{$\downarrow\downarrow$ Lower} Memory 
\\
\hline
\end{tabular}
\end{table*}

%%%GPT2 table
\begin{table*}[t]
\centering
\caption{Comparison of weight tying and clipping strategies under DP-SGD on SST-2 (GPT2). The best utility is achieved with untied weights, while the most efficient configuration in terms of memory is Ghost Clipping with untied weights.}
\label{tab:complete-dp_tradeoff-gpt2}

\renewcommand{\arraystretch}{1.3}
\scriptsize
\begin{tabular}{|p{3.6cm}|c|c|c|}
\hline
\textbf{Configuration} & \textbf{Accuracy (\%)} & \textbf{Memory (MB)}  &
\textbf{Effects}\\
\hline

\textbf{WT + Normal Clipping} 
& {78.44 $\pm$ 1.73\%}
& 18530.09 MB
& Baseline with moderate memory\\
\hline

\textbf{No WT + Normal Clipping} 
& \textbf{81.48 $\pm$ 1.07\%}
& 18973.67 MB 
& Untie weights $\Rightarrow$ \textcolor{green!45!black}{$\uparrow$ Higher} Utility \& \textcolor{red!45!black}{$\uparrow$ Higher} Memory\\
\hline

\textbf{No WT + Ghost Clipping} 
& {81.48 $\pm$ 1.07\%}
& \textbf{6786.90 MB} 
& Ghost clipping $\Rightarrow$ preserved Utility  \& \textcolor{green!45!black}{$\downarrow\downarrow$ Lower} Memory 
\\
\hline

\end{tabular}
\end{table*}
%____________________________

%%%

% \begin{figure}
% \centering
%   \includegraphics[width=0.5\textwidth]{table6-edited.png}
%   \caption{Complete DP perspective for GPT2 across weight tying and clipping paradigms. Untied embeddings consistently improve utility and stability under DP-SGD, while ghost clipping preserves these gains with lower memory usage. Under tied embeddings, however, ghost clipping requires explicit cross-term correction, restoring mathematical validity at the cost of significant computational overhead.}
% \label{fig:complete_gpt2_dp}
% \end{figure}

\subsection{Discussion}
Our findings reveal a previously overlooked tension between parameter-sharing architectural choices in decoder-only language models and scalable differentially private optimization.
%between modern LLM architectural design and scalable differentially private optimization%.
While ghost clipping was introduced as an efficient mechanism for reducing the computational burden of DP-SGD in Transformer-scale models, our analysis shows that its effectiveness depends on structural assumptions that are not universally satisfied by modern decoder-only architectures. In particular, the widespread adoption of weight tying in GPT-style models introduces shared parameter coordinates that violate the additive gradient separability assumption underlying ghost clipping.

A central insight of this work is that efficient DP mechanisms are not entirely architecture-agnostic. Existing ghost clipping formulations assume that parameter groups can be treated independently during norm decomposition. Under tied embeddings, however, the same parameter matrix simultaneously participates in both the input embedding and output projection pathways, producing coupled gradient contributions and an additional interaction cross-term in the true per-sample gradient norm. This changes the gradient geometry assumed by standard ghost norm computation and renders the conventional formulation incomplete for tied-weight decoder-only LLMs.

Our empirical results further demonstrate that the effect of weight tying differs substantially between private and non-private optimization regimes, as summarized in Table~\ref{tab:all_exp_comparison}. Under standard SGD, weight tying behaves as intended in prior language modeling literature: it preserves predictive utility while improving parameter efficiency and reducing memory consumption. In contrast, under DP-SGD, tied embeddings consistently reduce utility and increase optimization standard deviation across both DistilGPT2 and GPT2. Untying the embedding and output projection layers improves predictive accuracy while substantially reducing variability across random seeds. These observations suggest that the shared-gradient interactions introduced by weight tying become particularly problematic in the presence of clipping and noise injection, where the resulting gradient norms no longer conform to the assumptions underlying efficient norm computation.

More broadly, our results highlight the importance of jointly considering model architecture and privacy-preserving optimization algorithms. Many recent advances in scalable differential privacy focus primarily on improving the efficiency of clipping and accounting mechanisms, often assuming that architectural design choices are orthogonal to the privacy algorithm. Our findings demonstrate that this assumption does not always hold. Architectural mechanisms that are beneficial in conventional training may alter gradient structure in ways that affect both the correctness and efficiency of private optimization procedures.

To add, the trade-off exposed by our corrected Ghost formulation illustrates a broader challenge for private foundation-model training. Restoring mathematical correctness under tied embeddings is possible through explicit cross-term computation, but doing so substantially diminishes the computational advantages that motivate ghost clipping in the first place.

\begin{table*}[t]
\centering
\caption{Overall comparison of standard and differentially private training
under weight tying (WT) and untied embeddings (No-WT) on SST-2 For DistilGPT2 and GPT2.
Best DP utility is shown in bold.}
\label{tab:all_exp_comparison}

\renewcommand{\arraystretch}{1.15}
\setlength{\tabcolsep}{8pt}
\small

\begin{tabular}{l c cc cc}
\hline
\multirow{2}{*}{\textbf{Training Setting}}
& \multirow{2}{*}{\textbf{Embedding}}
& \multicolumn{2}{c}{\textbf{DistilGPT2}}
& \multicolumn{2}{c}{\textbf{GPT2}} \\
\cline{3-6}
&
& \textbf{Accuracy (\%)}
& \textbf{Std. Dev.}
& \textbf{Accuracy (\%)}
& \textbf{Std. Dev.} \\
\hline

\multirow{2}{*}{Standard SGD}
& WT
& 86.77 & 0.76
& 89.09 & 0.69 \\

& No-WT
& 86.77 & 0.76
& 89.09 & 0.69 \\
\hline

\multirow{2}{*}{DP-SGD (Normal)}
& WT
& 80.61 & 1.32
& 78.44 & 1.73 \\

& No-WT
& \textbf{83.23} & \textbf{0.25}
& \textbf{81.48} & \textbf{1.07} \\
\hline

DP-SGD (Ghost)
& No-WT
& \textbf{83.23} & \textbf{0.25}
& \textbf{81.48} & \textbf{1.07} \\
\hline

\multicolumn{2}{l}{\textit{Memory reduction: Ghost vs. Normal (No-WT)}}
& \multicolumn{2}{c}{\textbf{66.7\%}}
& \multicolumn{2}{c}{\textbf{64.3\%}} \\
\hline
\end{tabular}

\vspace{1mm}
\footnotesize
WT: weight tying; No-WT: untied input/output embeddings.
\end{table*}

Importantly, our experiments confirm that ghost clipping itself remains highly effective when its underlying assumptions are satisfied. Under untied embeddings, ghost clipping achieves identical utility and optimization stability to normal clipping under DP-SGD while reducing GPU memory consumption by more than 60\% across both models. These results demonstrate that the observed discrepancies do not stem from ghost clipping as an optimization technique, but rather from the interaction between its norm-decomposition assumptions and shared-parameter architectures. In other words, ghost clipping is exact when model parameters can be partitioned into disjoint coordinate blocks, but requires modification when architectural mechanisms such as weight tying introduce shared-parameter interactions.

Finally, our findings challenge the common assumption that architectural techniques beneficial in conventional deep learning remain equally advantageous under differential privacy. Weight tying has long been regarded as an effective design choice for decoder-only language models because it improves parameter efficiency without sacrificing predictive performance. Our results show that this intuition does not necessarily extend to private optimization. In the presence of clipping and noise injection, architectural parameter sharing can alter gradient structure in ways that conflict with the assumptions underlying efficient DP mechanisms. Consequently, future research on scalable private training should consider architectural design and privacy-preserving optimization as tightly coupled components rather than independent aspects of the learning pipeline.

%---------------------------------------------%
\section{Limitations and Future Work}
\label{sec:limitations_future_work}

While our work provides both empirical and theoretical evidence that standard ghost clipping assumptions break under weight tying in decoder-only LLMs, several limitations remain.

% First, our empirical evaluation is limited to GPT2 and DistilGPT2 due to the computational cost of DP-SGD. However, the identified cross-term arises from tied embeddings rather than model scale, suggesting that the same interaction should persist in larger decoder-only architectures. Evaluating this effect in billion-parameter models remains an important direction for future work.
First, our empirical evaluation is limited to GPT2 and DistilGPT2 and therefore does not establish that the observed utility effects extend to larger or more recent decoder-only models. However, the cross-term identified follows from shared parameter coordinates between the input embedding and output projection and is therefore not specific to GPT2 scale. Thus, while the mathematical incompatibility applies to architectures exhibiting the analyzed form of weight tying, the magnitude of its empirical utility and efficiency effects in larger models remains to be established.

Second, our analysis focuses on embedding weight tying. Modern foundation models employ other forms of parameter sharing, including shared experts and parameter-efficient adaptation techniques. Investigating whether similar violations of gradient separability arise in these settings is an interesting avenue for future research.

Finally, although the corrected norm formulation restores consistency under tied embeddings, computing the interaction term introduces substantial overhead. Developing efficient approximations or alternative clipping strategies that preserve correctness without sacrificing scalability remains an open challenge.

More broadly, our findings suggest that scalable DP optimization cannot always be designed independently of model architecture. Future private LLM systems may therefore require closer co-design between privacy mechanisms, optimization algorithms, and parameter-sharing strategies.

%------------------------------------------
\section{Conclusion}
\label{sec:conc}

This paper investigated the interaction between DP-SGD, ghost clipping, and weight tying in decoder-only language models. Through theoretical analysis and extensive experiments on GPT2 and DistilGPT2, we showed that the benefits of weight tying observed in standard non-private training do not naturally transfer to the differentially private setting. Across both models, untied embeddings consistently achieved superior privacy--utility--efficiency trade-offs, improving predictive performance and optimization stability while preserving the substantial memory savings provided by ghost clipping.

A key contribution is identifying a limitation of ghost clipping under tied embeddings. We show that weight tying introduces an interaction term in the per-sample gradient norm, violating the additive separability assumption used in ghost norm computation. %Consequently, the standard formulation becomes incomplete.%
Consequently, treating the two pathway contributions as independent parameter blocks omits their cross-term and does not recover the exact norm of the shared parameter.
Although exact norms can be restored by including the missing cross-term, this requires materializing gradient interactions, which ghost clipping aims to avoid.

Taken together, these findings demonstrate that scalable differentially private optimization is not fully architecture-agnostic. Its correctness depends on how architectural choices shape gradient structure. Ghost clipping remains exact and effective when parameters are disjoint, but tied embeddings violate its assumptions. This highlights the need to jointly design model architectures and privacy-preserving optimization methods for efficient private foundation models.\\

AI Disclosure: We used OpenAI ChatGPT to assist with language refinement, manuscript organization, and improvement of the presentation and clarity of selected sections of the paper. %All technical content, analysis, experiments, mathematical formulations, and interpretations were developed and verified by the authors with full responsibility. 

\section*{Appendix}
\label{sec: appendix}
This appendix reports additional QNLI and QQP datasets results that reinforce the observations presented in the main paper. 
%_____________________________
\subsection{Dataset Statistics}
\label{app:datasets}
Detailed statistics for all datasets used in this work, including task type, dataset size, and balanced training splits are reported in Table \ref{tab:glue-dataset-full}.

\begin{table}[h]
\centering
\caption{GLUE Dataset Statistics with Task Types and Balanced Training Sizes}
\label{tab:glue-dataset-full}
\footnotesize
\setlength{\tabcolsep}{3pt}

\begin{tabular}{|l|l|l|c|c|}
\hline
\textbf{Dataset} & \textbf{Task Type} & \textbf{Split} & \textbf{Size (MB)} & \textbf{\#Samples} \\
\hline

\multirow{3}{*}{SST-2}
& \multirow{3}{*}{Sentiment Analysis}
& Train      & 11.82 & 67,349 (59,560 bal.) \\
& & Validation & 0.15  & 872 \\
& & Test       & 0.32  & 1,821 \\
\hline

\multirow{3}{*}{QQP}
& \multirow{3}{*}{Paraphrase Detection}
& Train      & 63.85 & 363,846 (268,756 bal.) \\
& & Validation & 7.09  & 40,430 \\
& & Test       & 68.61 & 390,965 \\
\hline

\multirow{3}{*}{QNLI}
& \multirow{3}{*}{QA/NLI}
& Train      & 18.38 & 104,743 (104,732 bal.) \\
& & Validation & 0.96  & 5,463 \\
& & Test       & 0.96  & 5,463 \\
\hline
\end{tabular}
\end{table}

%______________________________
\subsection{Experimental Results of QNLI and QQP}
\label{sec: qnliandqqp}
The QNLI and QQP results further support the conclusions drawn from the SST-2 experiments. Across both datasets and model sizes, the configuration combining untied embeddings with Ghost Clipping consistently achieves the most favorable privacy--utility--efficiency trade-off. Compared to the baseline setting of weight tying with standard clipping, removing the weight-tying constraint yields improved predictive performance while Ghost Clipping substantially reduces GPU memory consumption. For example, on QNLI, DistilGPT2 improves from 72.59\% to 73.57\% accuracy while reducing memory usage from 15.2 GB to 8.1 GB as shown in Table \ref{tab:distilgpt2_dp_complete-QNLI}, and GPT2 improves from 63.95\% to 68.69\% accuracy while reducing memory from 20.0 GB to 11.6 GB as shown in Table \ref{tab:gpt2_dp_complete-QNLI}. Similar trends are observed on QQP in Tables \ref{tab:distilgpt2_dp_complete-QQP} and \ref{tab:gpt2_dp_complete-QQP}, where untied embeddings combined with Ghost Clipping maintain or improve utility while providing significant memory savings. These results demonstrate that the advantages of untied embeddings under DP-SGD are not specific to sentiment classification on SST-2, but generalize across multiple natural language understanding tasks. Consequently, these results provide additional empirical evidence that untied embeddings enable Ghost Clipping to operate under its intended assumptions while offering a more effective and scalable configuration for differentially private training of decoder-only language models.

%Tables
%% QNLI on DistilGPT2
\begin{table*}[!htbp]
\centering
\caption{Comparison of weight tying and clipping strategies under DP-SGD on QNLI (DistilGPT2). Untied embeddings improve utility, while Ghost Clipping preserves these gains with substantially lower memory usage.}
\label{tab:distilgpt2_dp_complete-QNLI}
\renewcommand{\arraystretch}{1.3}
\scriptsize
\begin{tabular}{|p{3.6cm}|c|c|c|c}
\hline
\textbf{Configuration} & \textbf{Accuracy (\%)} & \textbf{Memory (MB)}  &
\textbf{Effects}\\
\hline
\textbf{WT + Normal Clipping} 
& 72.59 $\pm$ 1.07
& 15157.99
& Baseline with moderate memory\\
\hline
\textbf{No WT + Normal Clipping} 
& \textbf{73.57 $\pm$ 0.69} 
& 15600.46 
& Untie weights $\Rightarrow$ \textcolor{green!45!black}{$\uparrow$ Higher} Utility \& \textcolor{red!45!black}{$\uparrow$ Higher} Memory\\
\hline
\textbf{No WT + Ghost Clipping} 
& \textbf{ 73.57 $\pm$ 0.69}
& \textbf{8068.70} 
& Ghost clipping $\Rightarrow$ preserved Utility \& \textcolor{green!45!black}{$\downarrow\downarrow$ Lower} Memory 
\\
\hline
\end{tabular}
\end{table*}

%% QNLI on GPT2
\begin{table*}[!htbp]
\centering
\caption{Comparison of weight tying and clipping strategies under DP-SGD on QNLI (GPT2). Untied embeddings improve utility, while Ghost Clipping preserves these gains with substantially lower memory usage.}
\label{tab:gpt2_dp_complete-QNLI}
\renewcommand{\arraystretch}{1.3}
\scriptsize
\begin{tabular}{|p{3.6cm}|c|c|c|}
\hline
\textbf{Configuration} & \textbf{Accuracy (\%)} & \textbf{Memory (MB)}  &
\textbf{Effects}\\
\hline
\textbf{WT + Normal Clipping} 
& 63.95 $\pm$ 0.81
& 20032.06
& Baseline with moderate memory\\
\hline
\textbf{No WT + Normal Clipping} 
& \textbf{ 68.69 $\pm$ 2.18} 
& 20474.68
& Untie weights $\Rightarrow$ \textcolor{green!45!black}{$\uparrow$ Higher} Utility \& \textcolor{red!45!black}{$\uparrow$ Higher} Memory\\
\hline
\textbf{No WT + Ghost Clipping} 
& \textbf{ 68.69 $\pm$ 2.18}
& \textbf{11641.93} 
& Ghost clipping $\Rightarrow$ preserved Utility \& \textcolor{green!45!black}{$\downarrow\downarrow$ Lower} Memory 
\\
\hline
\end{tabular}
\end{table*}

%QQP DistilGPT2
\begin{table*}[!htbp]
\centering
\caption{Comparison of weight tying and clipping strategies under DP-SGD on QQP (DistilGPT2). Untied embeddings improve utility, while Ghost Clipping preserves these gains with substantially lower memory usage.}
\label{tab:distilgpt2_dp_complete-QQP}
\renewcommand{\arraystretch}{1.3}
\scriptsize
\begin{tabular}{|p{3.6cm}|c|c|c|}
\hline
\textbf{Configuration} & \textbf{Accuracy (\%)} & \textbf{Memory (MB)}  &
\textbf{Effects}\\
\hline
\textbf{WT + Normal Clipping} 
& 71.88 $\pm$ 1.08
& 15158.75
& Baseline with moderate memory\\
\hline
\textbf{No WT + Normal Clipping} 
& \textbf{ 73.20 $\pm$ 1.05}
& 15601.22 
& Untie weights $\Rightarrow$ \textcolor{green!45!black}{$\uparrow$ Higher} Utility \& \textcolor{red!45!black}{$\uparrow$ Higher} Memory\\
\hline
\textbf{No WT + Ghost Clipping} 
& \textbf{ 73.20 $\pm$ 1.05}
& \textbf{8068.05} 
& Ghost clipping $\Rightarrow$ preserved Utility \& \textcolor{green!45!black}{$\downarrow\downarrow$ Lower} Memory 
\\
\hline
\end{tabular}
\end{table*}

%%QQP with GPT2
\begin{table*}
\centering
\caption{Comparison of weight tying and clipping strategies under DP-SGD on QQP (GPT2). Untied embeddings improve utility, while Ghost Clipping preserves these gains with substantially lower memory usage..}
\label{tab:gpt2_dp_complete-QQP}
\renewcommand{\arraystretch}{1.3}
\scriptsize
\begin{tabular}{|p{3.6cm}|c|c|c|}
\hline
\textbf{Configuration} & \textbf{Accuracy (\%)} & \textbf{Memory (MB)}  &
\textbf{Effects}\\
\hline
\textbf{WT + Normal Clipping} 
& 68.82 $\pm$ 1.19
& 20032.06 MB
& Baseline with moderate memory\\
\hline
\textbf{No WT + Normal Clipping} 
& \textbf{69.46 $\pm$ 0.63}
& 20474.68
& Untie weights $\Rightarrow$ \textcolor{green!45!black}{$\uparrow$ Higher} Utility \& \textcolor{red!45!black}{$\uparrow$ Higher} Memory\\
\hline
\textbf{No WT + Ghost Clipping} 
& \textbf{69.46 $\pm$ 0.63}
& \textbf{11641.39 MB} 
& Ghost clipping $\Rightarrow$ preserved Utility \& \textcolor{green!45!black}{$\downarrow\downarrow$ Lower} Memory 
\\
\hline
\end{tabular}
\end{table*}

\subsection{Sensitivity to the Privacy Budget for DistilGPT2 on SST-2}
\label{sec:epsilon_sensitivity}
To assess whether the observed effect of weight tying is specific to the
privacy budget used in the main experiments ($\epsilon=3$), we additionally
evaluate DistilGPT2 on SST-2 at $\epsilon \in \{1,5\}$ while keeping the remaining training configuration fixed. The results are reported in
Tables~\ref{tab:complete-dp_tradeoff-distligpt2-eps1} and~\ref{tab:complete-dp_tradeoff-distligpt2-eps5}.\\
The results show that the relative behavior of the evaluated configurations persists across privacy budgets. At $\epsilon=1$, untied embeddings improve
accuracy from 79.09\% with WT and normal clipping to 84.23\%, while at
$\epsilon=5$, accuracy improves from 79.56\% to 80.72\%. Moreover, at both
privacy budgets, Ghost Clipping with untied embeddings matches the utility
of normal clipping with untied embeddings while retaining substantially lower GPU memory consumption.\\
Together with the main results at $\epsilon=3$, these experiments indicate that the observed advantage of untying is not restricted to a single privacy budget. Across $\epsilon\in\{1,3,5\}$, untied embeddings consistently
outperform their tied counterparts under DP-SGD, with Ghost Clipping lowering memory.

% preserves the utility of the untied configuration with substantially lower
% memory requirements.

%_________________%
\begin{table*}[t]
\centering
\caption{Comparison of weight tying and clipping strategies under DP-SGD on SST-2 (DistilGPT2) with $\epsilon=1$. The best utility is achieved with untied weights, while the most efficient configuration in terms of memory is Ghost Clipping with untied weights.}
\label{tab:complete-dp_tradeoff-distligpt2-eps1}

\renewcommand{\arraystretch}{1.3}
\scriptsize
\begin{tabular}{|p{3.6cm}|c|c|c|}
\hline
\textbf{Configuration} & \textbf{Accuracy (\%)} & \textbf{Memory (MB)}  &
\textbf{Effects}\\
\hline

\textbf{WT + Normal Clipping} 
& 79.38 $\pm$ 0.41
& 13813.55
& Baseline with moderate memory\\
\hline

\textbf{No WT + Normal Clipping} 
& \textbf{82.25 $\pm$ 2.81} 
& 14256.41 
& Untie weights $\Rightarrow$ \textcolor{green!45!black}{$\uparrow$ Higher} Utility \& \textcolor{red!45!black}{$\uparrow$ Higher} Memory\\
\hline

\textbf{No WT + Ghost Clipping} 
& 82.25 $\pm$ 2.81
& \textbf{4747.19} 
& Ghost clipping $\Rightarrow$ preserved Utility  \& \textcolor{green!45!black}{$\downarrow\downarrow$ Lower} Memory 
\\
\hline
\end{tabular}
\end{table*}

%_________________%
\begin{table*}[t]
\centering
\caption{Comparison of weight tying and clipping strategies under DP-SGD on SST-2 (DistilGPT2) with $\epsilon=5$. The best utility is achieved with untied weights, while the most efficient configuration in terms of memory is Ghost Clipping with untied weights.}
\label{tab:complete-dp_tradeoff-distligpt2-eps5}

\renewcommand{\arraystretch}{1.3}
\scriptsize
\begin{tabular}{|p{3.6cm}|c|c|c|}
\hline
\textbf{Configuration} & \textbf{Accuracy (\%)} & \textbf{Memory (MB)}  &
\textbf{Effects}\\
\hline

\textbf{WT + Normal Clipping} 
& 80.14 $\pm$ 0.82
& 13813.55
& Baseline with moderate memory\\
\hline

\textbf{No WT + Normal Clipping} 
& \textbf{81.13 $\pm$ 0.58} 
& 14256.41 
& Untie weights $\Rightarrow$ \textcolor{green!45!black}{$\uparrow$ Higher} Utility \& \textcolor{red!45!black}{$\uparrow$ Higher} Memory\\
\hline

\textbf{No WT + Ghost Clipping} 
& 81.13 $\pm$ 0.58
& \textbf{4747.19} 
& Ghost clipping $\Rightarrow$ preserved Utility  \& \textcolor{green!45!black}{$\downarrow\downarrow$ Lower} Memory 
\\
\hline
\end{tabular}
\end{table*}

\end{document}